\documentclass{article}

\usepackage[preprint]{neurips_2025}

\usepackage[utf8]{inputenc} 
\usepackage[T1]{fontenc}    
\usepackage{hyperref}       
\usepackage{url}            
\usepackage{booktabs}       
\usepackage{amsfonts}       
\usepackage{wrapfig}
\usepackage{graphicx}       
\usepackage{subfigure}  
\usepackage{nicefrac}       
\usepackage{microtype}      
\usepackage{xcolor}         
\usepackage{algorithm}
\usepackage{algorithmic}
\usepackage{amsmath}
\usepackage{multirow}

\title{Spatial RoboGrasp: Generalized Robotic Grasping Control Policy}

\author{%
\begin{tabular}{c c c}
Yiqi Huang\thanks{Authors contributed equally.} & Travis Davies\textsuperscript{$\star$} & Jiahuan Yan \\
\textit{ZhiCheng AI} & \textit{ZhiCheng AI} & \textit{ZhiCheng AI} \\[2ex]
Jiankai Sun & Xiang Chen & Luhui Hu \\
\textit{Stanford University} & \textit{Peking University} & \textit{ZhiCheng AI} \\
\texttt{jksun@stanford.edu} & & \texttt{luhui@zhicheng-ai.com} \\[2ex]
\end{tabular}
}

\begin{document}

\maketitle

\begin{abstract}

   Achieving generalizable and precise robotic manipulation across diverse environments remains a critical challenge, largely due to limitations in spatial perception. While prior imitation-learning approaches have made progress, their reliance on raw RGB inputs and handcrafted features often leads to overfitting and poor 3D reasoning under varied lighting, occlusion, and object conditions. In this paper, we propose a unified framework that couples robust multimodal perception with reliable grasp prediction. Our architecture fuses domain-randomized augmentation, monocular depth estimation, and a depth-aware 6-DoF Grasp Prompt into a single spatial representation for downstream action planning. Conditioned on this encoding and a high-level task prompt, our diffusion-based policy yields precise action sequences, achieving up to 40\% improvement in grasp success and 45\% higher task success rates under environmental variation. These results demonstrate that spatially grounded perception, paired with diffusion-based imitation learning, offers a scalable and robust solution for general-purpose robotic grasping.
   
\end{abstract}

\section{Introduction}

Humans intuitively adapt their grasping strategies to novel objects and conditions. In contrast, enabling robots to replicate this innate ability, generalizing across diverse objects and dynamic environments, remains a crucial challenge. Recent advances in imitation learning, particularly diffusion-based policies \cite{diffusion-policy, dp, aloha}, have shown promise in modeling the complexity and multimodality of manipulation tasks. However, these models often rely heavily on raw RGB inputs and handcrafted features, making them vulnerable to overfitting and weak spatial reasoning—especially when confronted with novel object geometries, changing viewpoints, lighting variations, or occlusions \cite{grad-cam, li2023robustvisualimitationlearning}. Lacking explicit task grounding, such policies depend solely on data-driven patterns, which further limits their robustness and generalizability \cite{grad-cam}.

A major contributor to these limitations is the scarcity of perceptual diversity in existing datasets, which are often collected under controlled laboratory conditions \cite{lin2024data}. As a result, models trained in such environments tend to perform poorly when deployed in complex real-world settings. Meanwhile, autonomous driving research has demonstrated that spatial-temporal augmentations and multimodal perception strategies can significantly improve resilience to environmental variability \cite{self-driving, 3d-outset, night-haze}. We propose to bring similar robustness principles into robotic manipulation through enhanced perception and spatial guidance.

To this end, we introduce Spatial RoboGrasp, a unified framework designed to improve both generalization and precision by integrating domain-randomized image augmentation, monocular depth estimation, and structured 6-DoF Grasp Prompts. This multimodal perception module produces explicit spatial inputs that guide a diffusion-based policy toward accurate, contact-aware action generation—without the overhead of full point cloud processing. Conditioned on these enriched inputs and a high-level task prompt, our approach generates robust, goal-directed trajectories. We investigate the following research questions: (1) Can multimodal perception and spatial augmentations improve robustness under significant environmental variation? (2) Can explicit grasp affordances enhance few-shot generalization to novel objects? (3) How effectively can grasp-based spatial prompts guide policy behavior and improve task performance?

By bridging the gap between lab-controlled training and real-world deployment, this work builds on our previous efforts in grasp-guided imitation learning \cite{huang2025robograspuniversalgraspingpolicy} and robust multimodal visual perception \cite{davies2024spatiallyvisualperceptionendtoend}, integrating them into a unified spatial policy architecture for robotic manipulation in unstructured environments.

\begin{figure*}[ht]
    \centering
    \includegraphics[width=\textwidth]{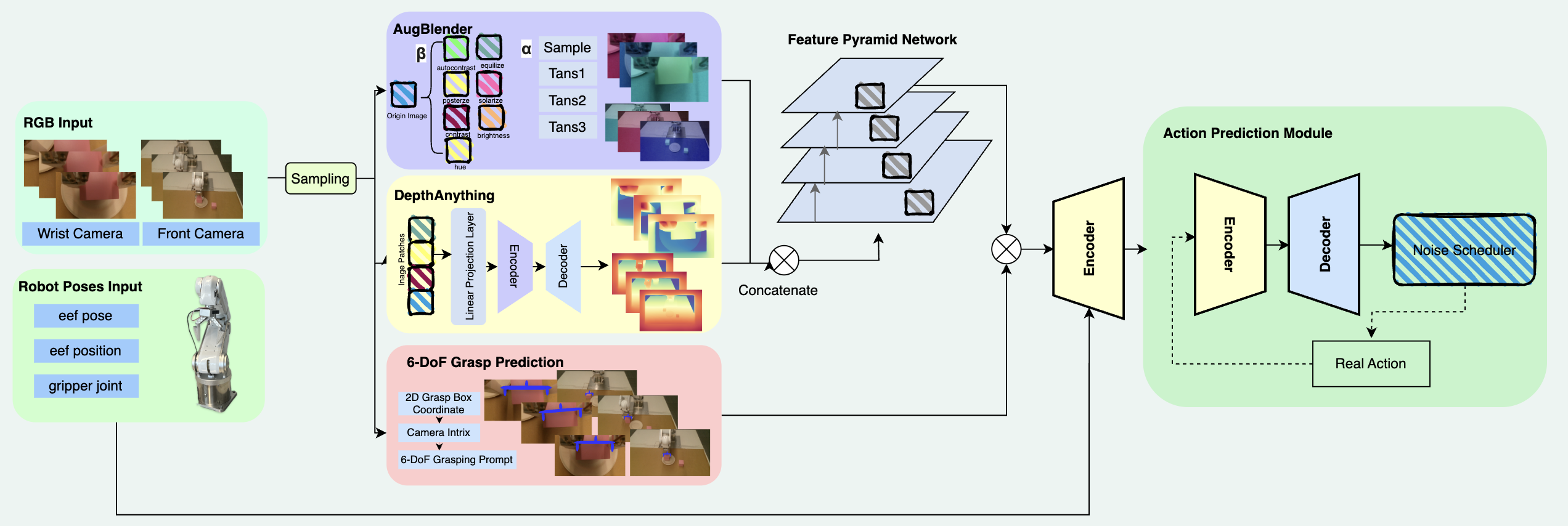}
    \caption{An overview Spatial RoboGrasp architecture, demonstrating the integration of image augmentations, depth estimation and a grasping prompt prediction modules. These observation conditions and robot state data to enhance generalizability and precision of grasping manipulation.}
    \label{fig:architecture}
\end{figure*}

\section{Related Works}

Recent advances in robot policy planning have broadened the scope of imitation learning (IL) beyond controlled lab environments \cite{aloha, aloha_2, umi}. IL frameworks typically map sensory observations directly to action sequences. Among them, diffusion-based policies such as Diffusion Policy (DP) \cite{diffusion-policy} have shown strong performance in tackling covariate shift—where a robot’s training distribution diverges from deployment settings \cite{covariate-shift, domain-generalisation}. These models generate diverse and multimodal trajectories, improving policy robustness in complex and uncertain environments.

The scalability of such IL methods has been further fueled by large-scale expert demonstration datasets \cite{x-embodiment}, enabling models like Robotics Diffusion Transformer (RDT) \cite{rdt} and $\pi_0$ \cite{pi0} to generalize to novel manipulation tasks with minimal supervision. CoinRobot \cite{zhao2025coinrobotgeneralizedendtoendrobotic} pushes this further by proposing a unified IL architecture that generalizes across robot embodiments and sensor configurations. However, despite this progress, their reliance on vast computational resources limits accessibility and practical deployment.

Beyond policy structure and data scale, real-world performance increasingly depends on robust spatial perception—especially in dynamic, visually diverse settings. Autonomous driving research highlights the importance of combining RGB data with spatial cues to maintain performance under adverse conditions \cite{radar-robust, hetero-av, robust-benchmark}. In robotics, spatial understanding has traditionally been enhanced through depth sensors, LiDAR, and multi-view 3D reconstruction \cite{rekep, cliport, okami2024, apollo, xpeng}. While effective, these approaches often incur high costs and demand precise calibration, limiting their feasibility for many applications.

Monocular depth estimation provides a promising alternative, enabling low-cost depth perception directly from RGB images \cite{midas, primedepth, depth-anything-v2}. Modern transformer-based models like Depth Anything V2 and DINOv2 \cite{depth-anything-v2, dinov2} offer real-time performance and robustness, making them well-suited for robotic deployment. However, their integration into IL pipelines remains limited, especially in settings that demand resilience to environmental variability. SVP \cite{davies2024spatiallyvisualperceptionendtoend} demonstrated that augmenting RGB data with monocular depth and domain-randomized corruptions (via AugBlender) improves model robustness across lighting conditions. Their findings show that structured visual perception can mitigate performance collapse under extreme exposure variation—a common limitation in RGB-only pipelines.

To address this gap, recent research emphasizes multimodal perception and data augmentation strategies. Domain-randomized augmentations such as AugMix \cite{augmix, self-driving, 3d-outset, night-haze} simulate realistic visual perturbations and, when combined with depth estimation, can help models learn more invariant spatial representations.

While robust perception is essential, manipulation performance ultimately hinges on accurate grasp planning. Affordance-based methods have emerged as a promising strategy by providing spatial priors that guide grasp selection \cite{grasping-survey}. Point-based affordances identify object locations but often lack sufficient detail for determining stable grasps \cite{moka, kalie, rekep}. In contrast, grasp-based affordances encode spatially grounded, actionable cues. Large-scale datasets like Grasp Anything \cite{grasp-anything} offer a path toward scalable training, but their integration with diffusion-based IL remains underexplored. Early efforts like GQCNN \cite{gqcnn} have demonstrated the potential of combining affordance reasoning with robotic control. Grasp-based affordances, such as those used in RoboGrasp \cite{huang2025robograspuniversalgraspingpolicy}, explicitly provide grasp poses or grasping boxes that enable consistent and precise manipulation. Their integration with diffusion policies improves few-shot and prompted grasp generalization, especially in tasks requiring spatial grounding.

These advances point to the need for a unified, lightweight framework that combines domain-randomized augmentation, monocular depth estimation, and spatially grounded grasp guidance. While each component has shown promise individually, their integration within a single pipeline for imitation learning remains underexplored. This gap motivates our proposed approach, which brings together these elements to enhance spatial reasoning and policy robustness in dynamic environments.

\section{Methodology}

Our approach addresses robust spatial perception and accurate grasping for robotic manipulation by integrating multimodal sensory information through a cohesive, structured pipeline. The architecture comprises four primary modules: AugFusion, Monocular Depth Estimation, Grasp Prediction, and a downstream Robotic Action Head as shown in Figure \ref{fig:architecture}. RGB image inputs are augmented to simulate realistic environmental variability, enriched with depth information for precise spatial awareness, and further enhanced with grasp predictions for explicit spatial affordances. These components collaboratively produce a rich and robust observation embedding, which the diffusion-based robotic action head leverages to generate stable, accurate manipulation policies.

\subsection{AugFusion}
To improve perception robustness under real-world visual variability, we introduce AugFusion, a domain-randomized augmentation strategy that extends AugMix \cite{augmix} by incorporating both in-distribution and out-of-distribution (OOD) RGB corruptions. Unlike AugMix, which focuses on mild, within-distribution transformations, AugFusion applies realistic corruptions such as lighting shifts, exposure changes, blur, and noise to simulate challenging conditions where RGB inputs degrade.

AugFusion uses a probabilistic mechanism controlled by parameter $\beta$ to decide between mixing multiple augmentations or applying them sequentially. Mixing weights are drawn from a Dirichlet distribution ($\alpha$), and a blending factor $\lambda$ adjusts intensity (see Algorithm~\ref{alg:AugFusion}). This process encourages the model to shift reliance toward depth cues when RGB becomes unreliable.

Integrating AugFusion-generated RGB data into our multimodal perception framework significantly broadens the model’s training distribution, enabling it to develop robust, invariant features essential for reliable real-world deployment under dynamic and unpredictable conditions.

\begin{algorithm}[H]

\begin{algorithmic}[1]
\REQUIRE Image $x$, number of chains/augmentations $k$, parameter $\alpha$, logic gate threshold $\beta$, mixing parameter $\lambda$
\STATE Randomly select $\xi \in [0,1]$
\STATE Mixing weights: $w \leftarrow \text{Dirichlet}(\alpha)$
\STATE Augmentations: $A \leftarrow \{a_1, \dots, a_n\}$
\STATE $\lambda \leftarrow \begin{cases} 1, & \text{if } \xi < \beta \\ \lambda, & \text{otherwise} \end{cases}$
\STATE $x_t \leftarrow x$
\FOR{$i$ in $\{1, \dots, k\}$}
    \STATE $x_{\text{aug}} \leftarrow x$
    \STATE Randomly select $a \subseteq A$ such that $|a| = k$
    \STATE Randomly select chain length $L \in \{1, \dots, k\}$
    \IF{$\xi > \beta$}
        \FOR{$a_i$ in $a \subseteq \{ a_1, \dots, a_L \}$} 
            \STATE $x_{\text{aug}} \leftarrow a_i(x_{\text{aug}})$
        \ENDFOR
        \STATE $x_t \leftarrow x_t + w_i \cdot x_{\text{aug}} $
    \ELSE
        \STATE $x_t \leftarrow a_i(x_t)$
    \ENDIF
\ENDFOR
\STATE $y \leftarrow \lambda \cdot x_{\text{t}} + (1 - \lambda) \cdot x$
\RETURN $y$

\caption{AugFusion}
\label{alg:AugFusion}
\end{algorithmic}
\end{algorithm}

\subsection{Monocular Depth Estimation Module}

Integrating depth information significantly enhances robustness by enriching our model's multimodal perception, allowing it to adaptively leverage complementary modalities during both training and inference. Depth data provides essential geometric context, which is especially beneficial for maintaining consistent performance under challenging perceptual variations, such as those caused by lighting changes. We specifically utilize monocular depth estimation due to its reliance solely on RGB images, enabling straightforward integration without the additional cost or complexity of dedicated hardware. Recent advances in monocular depth estimation models have demonstrated promising real-time capabilities combined with high accuracy and robustness, making them particularly suited to our robotic application \cite{midas, primedepth, depth-anything-v2}.

For our implementation, we adopt the Depth Anything V2 model, which leverages the transformer-based DINOv2 architecture explicitly optimized for robust monocular depth estimation \cite{dinov2}. Depth Anything V2 has shown impressive resilience to common image corruptions prevalent in robotic scenarios, aligning perfectly with our objective of achieving robust perception.

To enhance computational efficiency, we preprocess our entire training dataset by extracting depth maps from each RGB frame using the ViT-B-based variant of Depth Anything V2. This preprocessing step ensures tight alignment between RGB and depth modalities while significantly reducing memory consumption and accelerating the training pipeline. During inference, we employ the lighter ViT-S-based model variant, which maintains high-quality depth estimation at reduced computational costs, thereby achieving real-time inference suitable for deployment in resource-constrained robotic systems.

\subsection{Grasp Prediction Module}

\begin{wrapfigure}{l}{0.5\linewidth}  
    \centering
    \vspace{-1em}  
    \includegraphics[width=\linewidth]{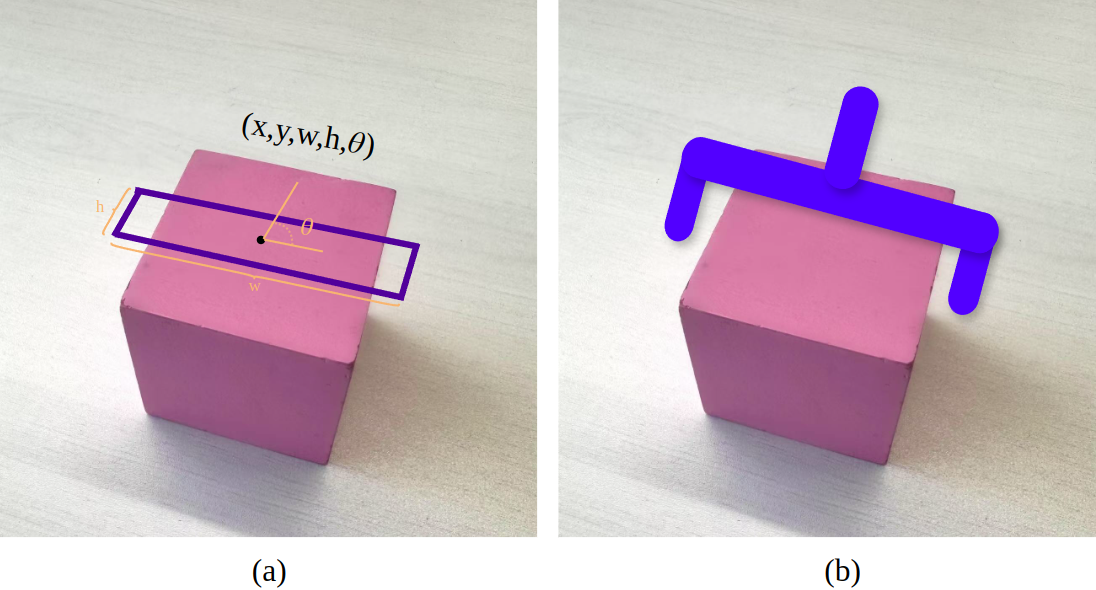}
    \caption{Grasp prompt visualization in the PickBig task. (a) Predicted oriented 2D grasp box from the RGB image. (b) Corresponding 6-DoF grasp prompt derived using camera intrinsics and a rotation matrix.}
    \label{fig:grasping_box}
    \vspace{-1em}
\end{wrapfigure}

To provide explicit grasp affordances for downstream manipulation, we employ a YOLO-based grasp detection module trained to predict 2D-oriented grasping boxes from RGB inputs. This module significantly extends grasping capability by leveraging rich depth-aware embeddings derived from the monocular depth estimation model, enabling accurate grasp localization without the computational overhead of explicit point-cloud generation. Specifically, we fine-tuned a lightweight YOLOv11-m model on a custom-labeled dataset containing annotated graspable regions, where each annotation consists of a 5D grasp representation (as shown in Figure \ref{fig:grasping_box}):
\[
(x, y, w, h, \theta)
\]
where $(x,y)$ is the center of the grasp box in image coordinates, $(w,h)$ denote its dimensions, and $\theta$ represents the in-plane rotation relative to the image frame.

While YOLO operates purely in 2D, we convert its predictions into full 6-degree-of-freedom (6-DoF) grasp poses through a lightweight, geometry-based postprocessing step. Given the estimated depth map from the monocular depth module and known camera intrinsics $(f_x, f_y, c_x, c_y)$, we project the 2D grasp center $(x, y)$ into 3D space as follows:
\[
x_{3D} = \frac{(x - c_x) \cdot z}{f_x}, \quad 
y_{3D} = \frac{(y - c_y) \cdot z}{f_y}, \quad 
z_{3D} = z
\]

where $f_x, f_y$ are focal lengths, $(c_x, c_y)$ the camera's principal point, and $z$ the depth sampled from the predicted depth map at pixel $(x, y)$.

To determine the rotation component of the grasp pose, we use the predicted in-plane angle $\theta$ to construct a rotation matrix, assuming a top-down grasp direction (aligned along the camera optical axis $Z$):
\[
R = [\mathbf{x}, \mathbf{y}, \mathbf{z}], \quad \text{where} \quad
\mathbf{x} = [\cos\theta, \sin\theta, 0], \quad 
\mathbf{z} = [0, 0, 1], \quad 
\mathbf{y} = \mathbf{z} \times \mathbf{x}
\]

The resulting rotation matrix $R$ can be converted to a quaternion representation $(q_x, q_y, q_z, q_w)$ for efficient use in downstream robotic planning.

Thus, the final predicted grasp pose is represented as:
\[
g = (x_{3D}, y_{3D}, z_{3D}, q_x, q_y, q_z, q_w)
\]

At both training and inference stages, the grasp detection module outputs a single high-confidence grasp box per frame. During inference, this 2D box—comprising its center coordinates, size, and in-plane angle—is transformed into a full 6-DoF pose using the predicted depth map and known camera intrinsics.

We define this 6-DoF grasp pose as a 6-DoF Grasp Prompt: a predicted end-effector pose and gripper width derived from RGB-D input, which explicitly conditions the diffusion policy to guide contact-aware grasping decisions. This grasp prompt is passed to the observation encoder as an explicit, spatially grounded condition. By leveraging a fast, pretrained YOLOv11 model, our design retains real-time speed and data efficiency while avoiding the overhead of point cloud processing or retraining complex grasp regressors.

Furthermore, this setup remains fully compatible with existing grasp annotations and seamlessly integrates into our multimodal pipeline. The resulting 6-DoF grasp predictions provide actionable, contact-aware guidance to the diffusion-based policy, enhancing spatial reasoning, generalization, and manipulation precision in diverse, real-world environments.

\subsection{Observation Encoder}
\begin{wrapfigure}{r}{0.5\linewidth}  
    \centering
    \vspace{-1em}  
    \includegraphics[width=\linewidth]{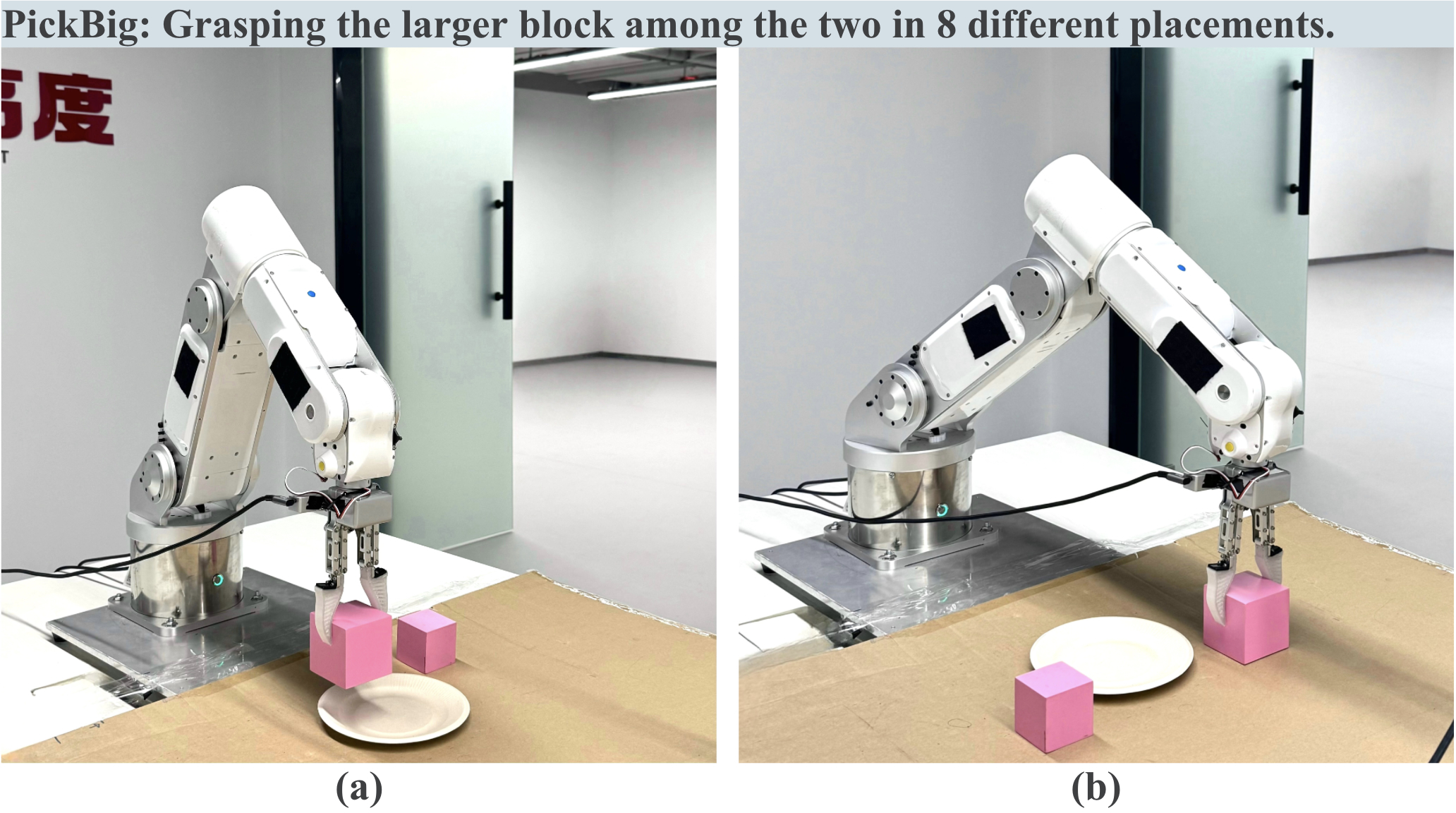}
    \caption{Placement generalization setup for the PickBig task. (a) and (b) illustrate two of the eight object configurations. The goal is to identify and grasp the larger of two similarly shaped blocks along its diameter.}
    \label{fig:pickbig_arm}
    \vspace{-1em}
\end{wrapfigure}
The observation encoder fuses multiview RGB inputs, low-dimensional robot states, and grasp-specific features into a unified spatial-temporal representation for diffusion-based policy learning. Each fixed camera view is processed independently by a ResNet-34-based Feature Pyramid Network (FPN) \cite{fpn, resnet}, enabling multi-scale feature extraction tailored to different viewpoints. Features from all views are pooled and concatenated into a comprehensive visual embedding.

This representation is augmented with AugFusion-processed RGBs, monocular depth maps, and 6-DoF grasp poses predicted by the grasp detection module. Additionally, robot end-effector state, gripper status, and task prompt embeddings are concatenated and linearly projected into a fixed-dimensional token per timestep.

A lightweight transformer then applies self-attention over tokens from the previous two timesteps, producing a spatially grounded, temporally aware encoding to condition the diffusion policy

\subsection{Robotic Action Head}
\begin{wrapfigure}{l}{0.5\linewidth}  
    \centering
    \vspace{-1em}  
    \includegraphics[width=\linewidth]{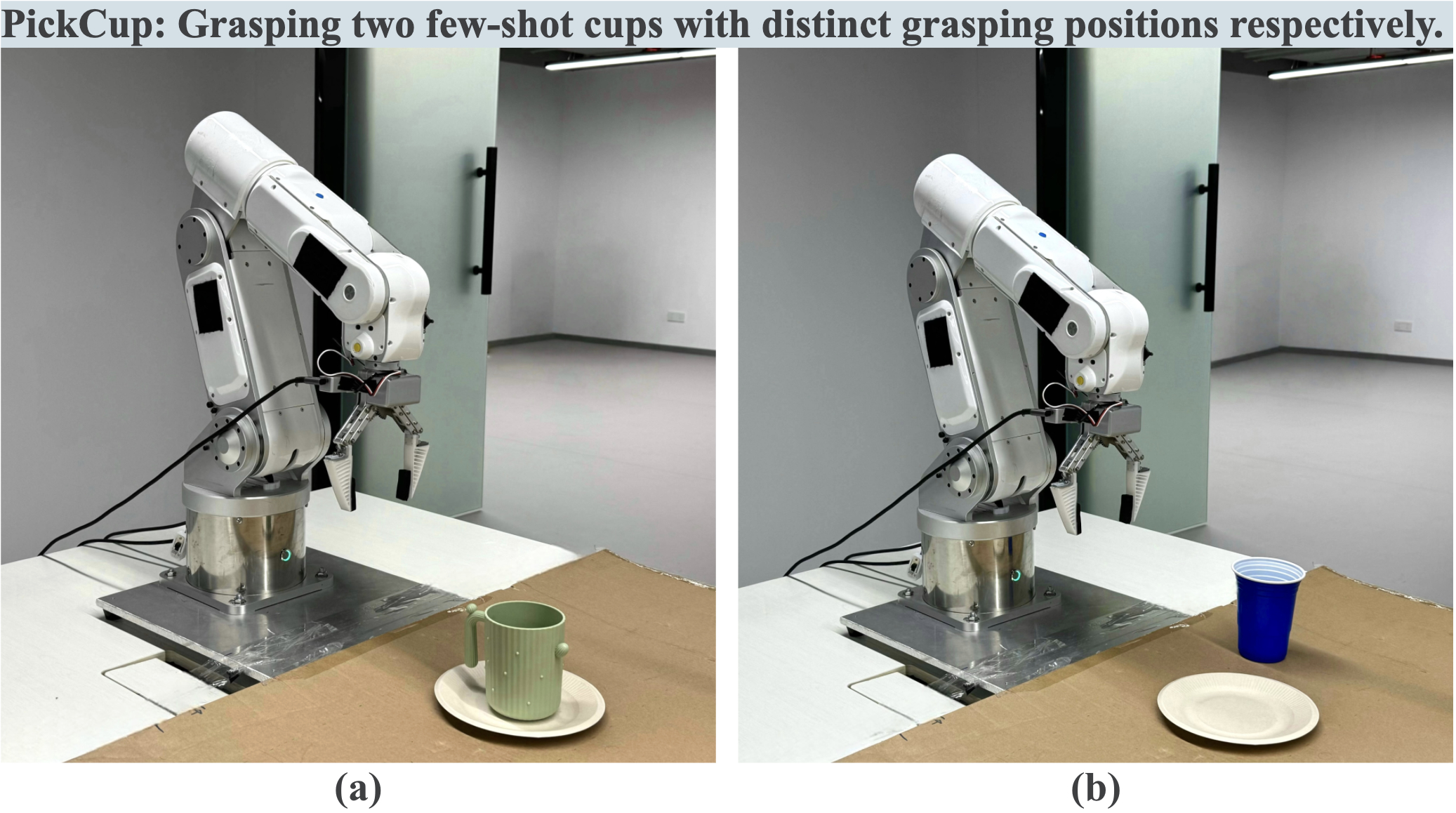}
    \caption{Few-shot PickCup setup. (a) Green mug with 5-shot handle grasping. (b) Blue plastic cup with 10-shot diameter grasping.}
    \label{fig:pickCup_arm}
    \vspace{-1em}
\end{wrapfigure}
Our action head adopts a diffusion-based policy \cite{dp}, which models action generation as a denoising process from Gaussian noise to expert trajectories. Instead of direct regression, it refines sampled noise over 16 timesteps using a DDIM scheduler \cite{iddpm} with a cosine beta schedule, capturing multimodal and stochastic dynamics.

A transformer with cross-attention layers conditions the denoising on observation tokens. Actions are projected into a latent space, iteratively refined, and reprojected back via a linear head. Training uses Root Mean Square Error (RMSE) loss for its precision sensitivity, enabling the model to produce smooth and goal-aligned action sequences from multimodal inputs.

\section{Experiments and Evaluation}

Robotic manipulation in real-world settings demands precise grasping and strong generalization across diverse objects and conditions, yet prior imitation learning studies often rely on limited, controlled lab setups. To bridge this gap, we introduce a comprehensive experimental suite—PickBig, PickCup, and PickGoods—that evaluates generalization across object scales, categories, grasp strategies, and few-shot or prompt-driven scenarios. Our ablation studies analyze the impact of grasp affordances, monocular depth, and augmentation, while controlled exposure tests benchmark visual robustness under realistic lighting variation.

\subsection{Task Description}

\textbf{PickBig}:
This task tests the robot’s ability to select and grasp the larger of two visually similar blocks placed in eight spatial configurations. The challenge lies in discerning subtle size differences and adapting the grasp accordingly (see Figure \ref{fig:pickbig_arm}). 

\textbf{PickCup}:
This task evaluates generalization across diverse cup geometries and grasp types—handle, wall, and diameter grasps—across four positions. Few-shot trials with new cup types assess the policy’s ability to extend learned strategies with minimal data, guided by 6-DoF grasp representations (see Figure \ref{fig:pickCup_arm}).

\textbf{PickGoods}:
This task simulates open-world manipulation using varied consumer items. Each object is associated with a fixed grasp strategy, and a grasping prompt defines the target. This tests whether the policy can execute goal-directed actions when spatial intent is made explicit (see Figure \ref{fig:pickGoods_arm}).

\begin{itemize}
    \item PickBig: 600 trials across 8 placement configurations, distinguishing and grasping the larger of two similar blocks.
    \item PickCup: 315 trials covering three cup types with varying grasping strategies (handle, sidewall, diameter), plus 15 few-shot examples for underrepresented cups.
    \item PickGoods: 400 trials involving four retail items, each paired with a consistent grasp strategy to evaluate promptable grasping.
\end{itemize}
\subsection{Data Processing}

We curate task-specific datasets to support robust and generalizable grasp policy learning:
\begin{wrapfigure}{r}{0.5\linewidth}  
    \centering
    \vspace{-1em}  
    \includegraphics[width=\linewidth]{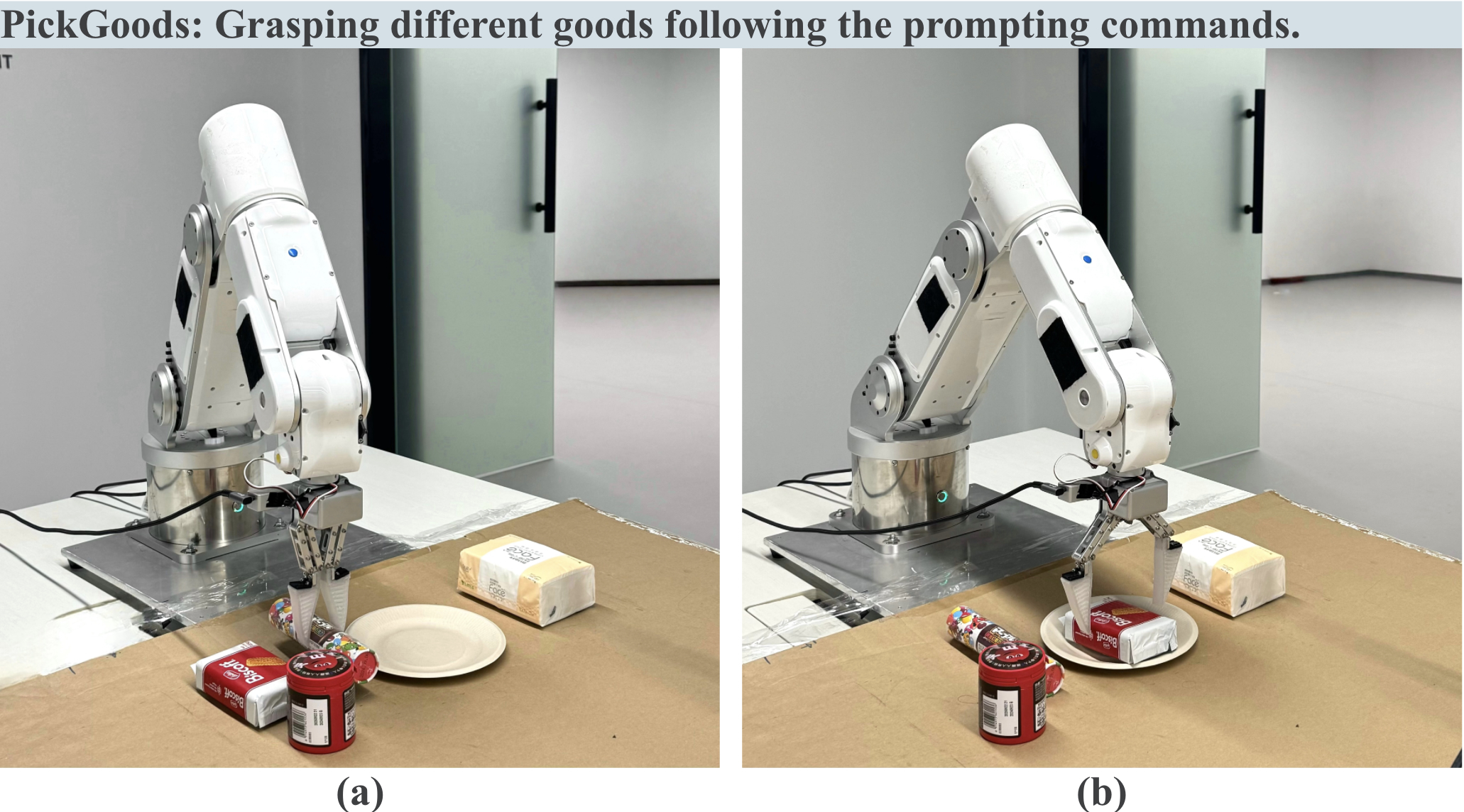}
    \caption{Promptable PickGoods setup. (a) Grasp prompt for a chocolate bar; (b) for a biscuit. The goal is to follow prompts to pick the target item.}
    \label{fig:pickGoods_arm}
    \vspace{-1em}
\end{wrapfigure}
Each demonstration includes synchronized RGB and monocular depth frames. A representative subset is annotated with 2D grasp boxes, used to fine-tune our grasp detector. These 2D grasp annotations are then converted into 6-DoF grasp poses using depth and fixed camera intrinsics—projecting 2D centers to 3D, estimating orientation from in-plane angles, and deriving gripper widths from box dimensions.

This automated grasp pipeline generates consistent 6-DoF annotations for the full dataset and enables real-time grasp prediction at inference. By eliminating the need for point cloud reconstruction, it supports scalable, accurate, and contact-aware grasp supervision integrated directly into our multimodal observation encoder.

\subsection{Evaluation Metrics}
To assess both task-level performance and environmental robustness, we employ a set of complementary metrics across all experiments. Specifically, we evaluate:

Task Success Rate (TSR): The percentage of successful task completions, measuring whether the robot accomplishes the overall objective.

Grasp Success Rate (GSR): The proportion of stable and accurate grasp executions, capturing action consistency and spatial precision. It is computed as:

\begin{equation}
GSR = \frac{\text{No. Successful Grasps}}{\text{No. Total Grasp Attempts}}
\end{equation}
 
To rigorously evaluate robustness under lighting variation, we conduct controlled tests across 10 discrete camera exposure levels (10 to 170 ms). For each exposure setting, every model is evaluated over 100-150 trials per task. Success is determined through consensus among 2–3 human evaluators, who assess whether the manipulation was executed correctly. Final scores are computed by averaging success rates across all exposure levels, providing a holistic view of the model's adaptability and reliability in realistic, visually corrupted conditions.

\subsection{Computation Resources}
Our setup used a standard industrial robot arm with two RGB cameras: one front-facing and one wrist-mounted for close-up views. As most imitation learning methods are designed for simulation, we selected Diffusion Policy (DP) as our baseline due to its proven effectiveness in real-world settings. We retrained DP with extensive hyperparameter tuning and applied our improvements under similar configurations. Given the unreliability of loss curves in robotic learning, we trained each model for a fixed duration based on expert heuristics. Training typically took around 48 hours per model using a dual RTX 4090 GPU server.

\section{Results and Discussion}
Table \ref{tab:tsr_gsr_multirow} summarizes TSR and GSR across models and exposure levels. Our full model consistently outperforms all baselines across tasks and lighting conditions, validating the synergy of 6-DoF grasp affordances, monocular depth estimation, and AugFusion augmentation.

Each module contributes complementary benefits: depth alone boosts PickBig TSR by 34\%, highlighting the value of geometric priors; AugFusion improves PickCup TSR by 24\% under mid-range exposures; GraspPrompt enhances spatial precision in PickBig and PickGoods but struggles in clutter. When combined, these components yield the most robust performance, outperforming baselines by 15–30\% across metrics.

\subsection{Lighting Robustness Across Exposure Conditions}

Our ablation study reveals the limitations of existing diffusion policy variants when operating under environmental variability. Across all tasks, baseline DP and its data-augmented variants (e.g., DP+Depth, DP+AugFusion) exhibit steep performance declines under extreme lighting—particularly at low (10–40 ms) and high (160–170 ms) exposures. In contrast, our method maintains high TSR and GSR across the full exposure range, peaking around natural lighting (80–120 ms) and degrading gracefully at extremes. This is a direct result of our structured perception pipeline and its robustness to visual corruptions.

For instance, in the PickBig task, our method achieves an average TSR of 82\% and GSR of 81\%, outperforming all four baselines by over 14\% in both metrics. In PickCup, where lighting variation strongly affects visual cues for grasping handles or rims, our model sustains 82\% TSR and 80\% GSR, compared to DP’s 27\% and 21\%, respectively. This confirms our approach’s resilience to appearance changes through both visual and geometric embeddings.

\subsection{Grasp Precision and Strategy Learning}

The inclusion of GraspPrompts significantly improves performance across all grasping strategies. In PickCup, our model reliably distinguishes handle, wall, and diameter grasps with minimal demonstrations—even generalizing to unseen cups in few-shot settings. The prompt-driven PickGoods task further highlights our model’s capacity to attend to spatial objectives: while DP and other baselines perform near-zero due to ambiguity in cluttered scenes, our model reaches a 47\% TSR and 44\% GSR despite spatial complexity and distractors. This demonstrates the value of explicit spatial prompting for policy grounding.

\begin{table*}[ht]
\centering
\footnotesize 
\setlength{\tabcolsep}{5pt}  
\caption{Success rates (\%) for Task Success Rate (TSR) and Grasp Success Rate (GSR) across exposure levels. Each model spans two rows. AVG is the mean across all exposures. Bold font represents our better performance.}
\label{tab:tsr_gsr_multirow}
\begin{tabular}{lll|cccccccccc|c}
\toprule
\textbf{Task} & \textbf{Model} & \textbf{Metrics} & \textbf{10} & \textbf{20} & \textbf{40} & \textbf{60} & \textbf{80} & \textbf{100} & \textbf{120} & \textbf{140} & \textbf{160} & \textbf{170} & \textbf{AVG} \\
\midrule

\multirow{10}{*}{PickBig} 
 & DP              & TSR &  9  &  19   &  27   &  53   &  59   &   87  &   68 & 39    &  33   &   28 &  42\\
 &                 & GSR &  0   &  12   &   25  &   52  &   56  &  86   &   66  &  38   &  32   &  21   &  39\\
 & +Depth          & TSR &  53   &   82  &   78  &  75   &   83  &   82  &  90   &   75  &  70   &   68  &  76\\
 &                 & GSR &   53  &   80  &  77   &  73   &   83  &  80   &   87  &  73   &   67  &  65  & 74\\
 & +AugFusion     & TSR &  51   &  65   &  72  &   75  &  80   &   89  &  51   &  61   &  60   &   58  &  67 \\
 &                 & GSR &  49   &  62   &   70 &   73  &  78   &  87   &  49   &  59   &  58   &  56   & 64\\
 & +GraspPrompt    & TSR &   52  &   63  &  78   &  85   &  87   &  94   &   98   &  77   &  54   &  31 &  72\\
 &                 & GSR &   51  &  60   &   75  &  84   &   83  &  94   &   96  &   72  &  51   &   29  &  70\\
 & \textbf{Ours}   & TSR &  \textbf{62}   &   \textbf{65}  &  \textbf{76}   &  \textbf{89}   &  \textbf{91}   &  \textbf{95}   &   \textbf{98}  & \textbf{93}    &   \textbf{82}  &   \textbf{72}  &  \textbf{82}\\
 &                 & GSR &   \textbf{59}  &   \textbf{65}  &   \textbf{72}  &  \textbf{88}   &   \textbf{91}  &  \textbf{95}   &   \textbf{98}  &  \textbf{91}   &   \textbf{80}  &   \textbf{67} & \textbf{81}\\
                    
\midrule
         
\multirow{10}{*}{PickCup} 
 & DP              & TSR &  0   &  0   &   0  &  53   &   75  &   82  &   40  &  17   &  3   &   0 & 27\\
 &                 & GSR &  0   &  0   &   0  &   28  &   64  &   70  &  36   &   8  &  0   &   0 & 21\\
 & +Depth          & TSR &  0   &   32  &  60   &  67   &  80   &  87   &  89  &  75   &   53  &   26   &  57  \\
 &                 & GSR &  0   &   27  &   41  &  55   &  64   &  76  &   80  &   71  &   48  &   22  &  48  \\
 & +AugFusion     & TSR &  0   &   32  &  50   &  57   &  64   &  76   &  81  &  69   &   53  &   26   &  51  \\
 &                 & GSR &  0   &   15  &   32  &   50  &  56  &   70  &  77   &   61  &  33   &   17  &  41 \\
 & +GraspPrompt    & TSR &  0   &  37   &  54   &  77   &  85   &   98  &  98   &  79  &   55  &   0  &  58 \\
 &                 & GSR &  0   &  35   &  52  &  73   &  82   &   98  &  98   &   70  &   49  &   0  &  56  \\
 & \textbf{Ours}   & TSR &   \textbf{61}  &  \textbf{72}   &  \textbf{86}   &  \textbf{87}   &  \textbf{93}   &  \textbf{97}   &   \textbf{99}  &  \textbf{85}   &  \textbf{78}   &   \textbf{62}  & \textbf{82}  \\
 &                 & GSR &    \textbf{60} &   \textbf{69}  &  \textbf{85}   &  \textbf{85}   &   \textbf{93}  &  \textbf{96}  &  \textbf{99}   &   \textbf{84}  &   \textbf{73}  &  \textbf{60}   &  \textbf{80}  \\

\midrule

\multirow{10}{*}{PickGoods} 
 & DP              & TSR &   0  &   0  &  0   &   0  &  9   &   17  &  22   &  20   &  0   &  0  & 7\\
 &                 & GSR &  0   &   0  &  0   &   0  &   8  &  8   &   22  &   10  &  0   &   0 & 5\\
 & +Depth          & TSR &  0   &  0   &   0  &  16   &  25   &  34   &  41   &   28  &  0   &   0  & 14\\
 &                 & GSR &  0   &  0   &   0  &   13  &   20  &   25  &  27   &  18   &  0   &   0 & 10\\
 & +AugFusion     & TSR &   0  &   0  &   0  &   7  &   19  &  21   &   11  &  10   &   0  &  0  & 7\\
 &                 & GSR &   0  &   0  &   0  &   3  &   8  &   14  &   5  &   2  &  0   &   0 & 3\\
 & +GraspPrompt    & TSR &   0  &   0  &  0   &  19   &  31   &  49   &  52   &   36  &   12  &   0  &  20\\
 &                 & GSR &  0   &   0  &  0   &   9  &   25  &  29   &  51  &   32  &   7  &  0  &  15\\
 & \textbf{Ours}   & TSR &  \textbf{26}   &  \textbf{33}   &   \textbf{48}  &  \textbf{52}   &   \textbf{57}  &  \textbf{63}   &  \textbf{65}   &  \textbf{51}   &  \textbf{44}   &  \textbf{29}   & \textbf{47} \\
 &                 & GSR &  \textbf{22}  &   \textbf{28}  &   \textbf{43}  &   \textbf{50}  &  \textbf{56}   &   \textbf{61}  &   \textbf{64}  &   \textbf{51}  &   \textbf{40}  &  \textbf{21}   &  \textbf{44}  \\

\bottomrule
\end{tabular}
\end{table*}

\subsection{Discussion on Generalization, Few-Shot, and Promptability}

\textbf{Generalization to Diverse Objects:} Our results in PickBig show how spatial variation challenges policies relying on implicit low-dim state. Our model’s grasp predictions provide spatial priors that allow robust re-localization and differentiation, even for near-identical objects.

\textbf{Few-shot Transfer:} In PickCup, 5–10 demo generalization to new objects shows the model can abstract strategy from shape and appearance when supported by geometric priors.

\textbf{Prompt-following:} PickGoods highlights that spatial cues alone are insufficient without environmental diversity in training. Our model still succeeds by disambiguating goal directionality (e.g., chocolate vs. tissue), but future work should incorporate more clutter-aware prompting and pose diversity.

Our results demonstrate that combining structured 6-DoF grasp prompts, monocular depth, and robust visual augmentation enables a powerful spatial perception stack for generalizable and precise manipulation. The proposed approach consistently outperforms strong baselines in both task success and grasp execution, especially under environmental variation. This affirms our core claim: robust robotic grasping requires not just diverse inputs, but structured, spatially-grounded ones.

\section{Conclusion}

We present a unified framework that enhances both the spatial robustness and grasping generalization of robotic learning systems, enabling precise, contact-aware manipulation under diverse environmental conditions. By integrating AugFusion, monocular depth estimation, and 6-DoF grasp prediction into a cohesive perception module, our system significantly improves visuomotor policy performance without the need for additional 3D sensing hardware. Built atop a diffusion-based action model, this approach proves effective across a variety of tasks, object types, and lighting scenarios.

Through comprehensive experiments and ablation studies, we demonstrate that each module—depth, augmentation, and grasp prompt—contributes uniquely to policy robustness. Our full model achieves the highest task and grasp success rates across all exposure levels, showcasing strong few-shot generalization and reliable performance in unstructured, low-visibility environments. The results underscore the importance of spatially grounded perception in improving goal alignment, action precision, and environmental adaptability.

This work provides a scalable, plug-and-play solution for general-purpose robotic manipulation, bridging the gap between controlled lab setups and the complexities of real-world deployment.

\section{Future Work}

Building on our findings, future efforts should prioritize standardized benchmarks for imitation learning under real-world variability, similar to ImageNet-C or 4Seasons in computer vision. These would enable consistent evaluation of robustness and generalization across robotic systems. To streamline the grasping pipeline, an MLP grasp predictor could be trained end-to-end alongside depth and AugFusion features, replacing the standalone YOLO module. This approach may improve learning efficiency and offer tighter integration between visual cues and grasp prediction, while still supporting full 6-DoF outputs.

Improving temporal reasoning remains an open challenge. Our method processes short sequences but lacks long-horizon memory. Integrating memory-augmented architectures—such as Scene Memory Transformer or REMEMBR—could support policy continuity in dynamic or delayed-reward scenarios. Finally, extending our framework with language-conditioned prompting (e.g., via Grounding DINO or DINO-X) and applying grasp affordance prompts in world models or large-scale policy frameworks like ACT or Robotics Diffusion Transformer would further enhance task generality and semantic grounding. These directions aim to bridge spatial reasoning, multimodal inputs, and scalable policy learning in real-world environments.

\begingroup
\small
\bibliographystyle{plainnat}  
\bibliography{main}
\endgroup

\clearpage
\appendix

\section{Technical Appendices and Supplementary Material}

\begin{figure*}[ht]
    \centering
    \includegraphics[width=0.6\textwidth]{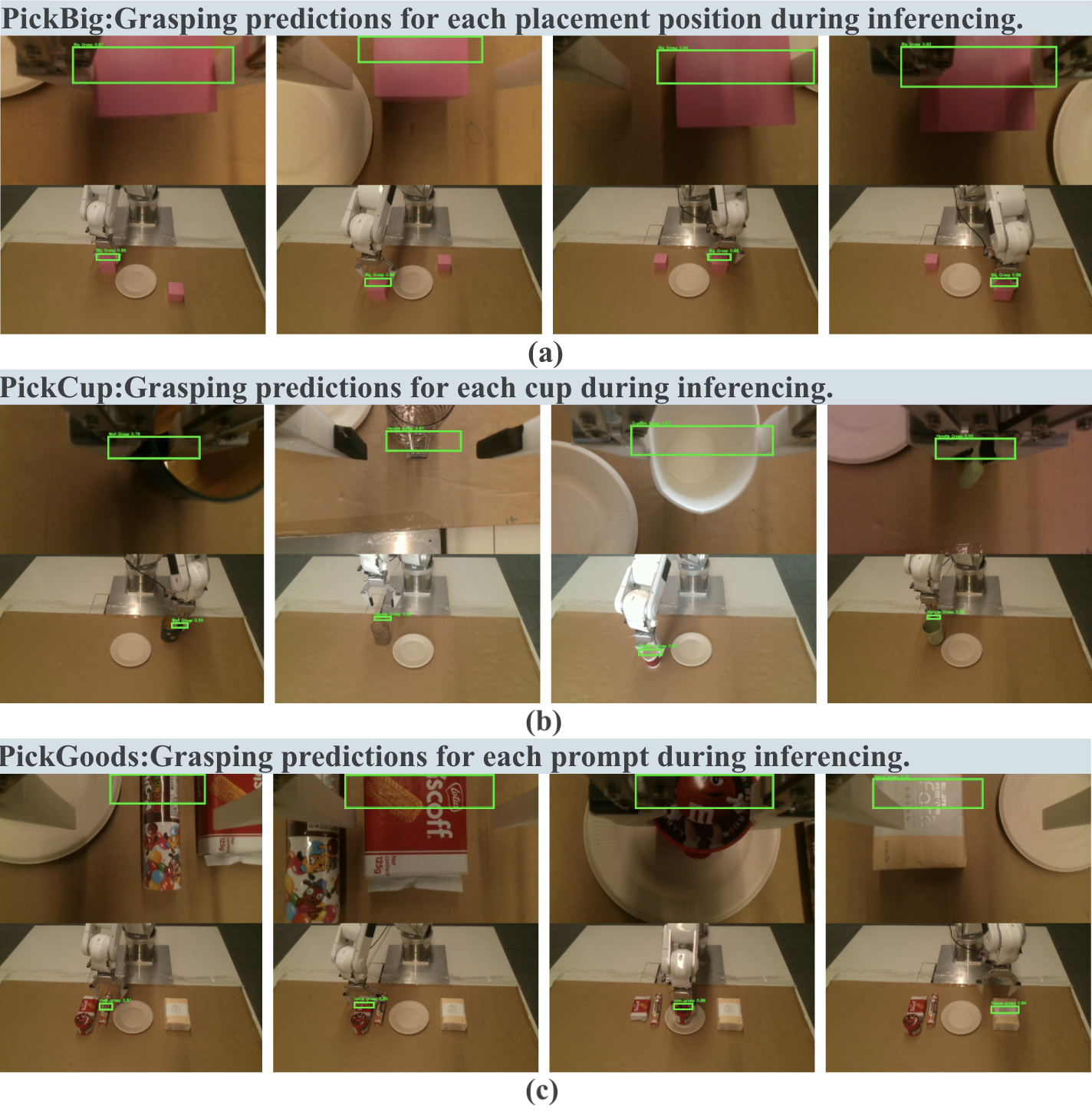}
    \caption{Real-time orientated-2D grasp box predictions across tasks using Spatial RoboGrasp. It will be later converted to 6-DoF grasp prompts:
(a) Robust grasp detection across varied placement configurations in PickBig.
(b) Accurate strategy-specific predictions for diverse cup geometries in PickCup.
(c) Prompt-guided grasp localization for goal-driven manipulation in PickGoods.}
    \label{fig:yolo_predict}
\end{figure*}

\begin{figure*}[ht]
    \centering
    \includegraphics[width=0.6\textwidth]{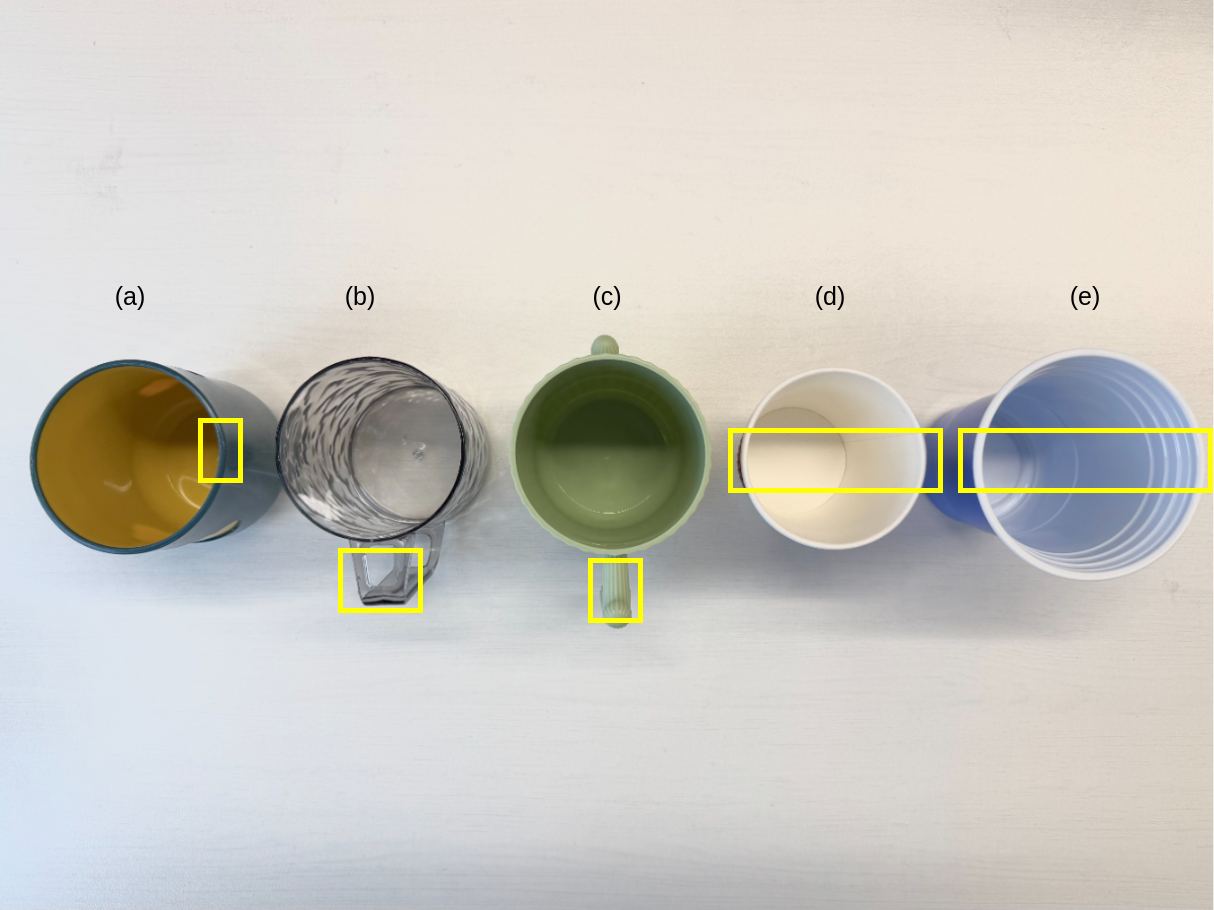}
    \caption{Illustration of grasping strategies for the PickCup task (top-down view).
(a) shows a sidewall grasp; (b) and (c) illustrate handle grasps; (d) and (e) depict diameter grasps. (c) and (e) correspond to the cups used in the few-shot evaluation.}
    \label{fig:pickCup_pos}
\end{figure*}


\end{document}